\documentclass[10pt,twocolumn,letterpaper]{article}

\usepackage{cvpr}
\usepackage{times}
\usepackage{epsfig}
\usepackage{graphicx}
\usepackage{amsmath}
\usepackage{amssymb}

\usepackage{graphicx}
\usepackage{graphbox}
\usepackage{gensymb}
\usepackage{float}
\usepackage{multirow}
\usepackage{booktabs}
\usepackage{ifthen}
\usepackage{array}
\usepackage{xcolor}

\newcommand*{\todo}[2][]{\textcolor{red}{[\textbf{\ifthenelse{\equal{#1}{}}{TODO}{TODO(#1)}}: #2]}}

\newcommand*\samethanks[1][\value{footnote}]{\footnotemark[#1]}

\DeclareMathOperator*{\argmax}{arg\,max}

\newcolumntype{L}[1]{>{\raggedright\let\newline\\\arraybackslash\hspace{0pt}}m{#1}}
\newcolumntype{C}[1]{>{\centering\let\newline\\\arraybackslash\hspace{0pt}}m{#1}}
\newcolumntype{R}[1]{>{\raggedleft\let\newline\\\arraybackslash\hspace{0pt}}m{#1}}

\definecolor{legendgreen}{RGB}{39, 174, 96}
\definecolor{legendblack}{RGB}{0, 0, 0}
\definecolor{legendred}{RGB}{211, 47, 47}

\newcommand*\diff{\mathop{}\!\mathrm{d}}

\cvprfinalcopy 

\ifcvprfinal
\usepackage[breaklinks=true,bookmarks=false]{hyperref}
\else
\usepackage[pagebackref=true,breaklinks=true,letterpaper=true,colorlinks,bookmarks=false]{hyperref}
\fi

\def\cvprPaperID{4251} 

\ifcvprfinal\fi
\begin{document}

\title{Query-efficient Black-box Adversarial Examples}

\author{
Andrew Ilyas\thanks{Equal contribution}\ \ $^{12}$,
\ Logan Engstrom\samethanks\ \ $^{12}$,
\ Anish Athalye\samethanks\ \ $^{12}$,
\ Jessy Lin\samethanks\ \ $^{12}$ \\
$^{1}$Massachusetts Institute of Technology, $^2$LabSix \\
\texttt{\{ailyas,engstrom,aathalye,lnj\}@mit.edu}
}

\maketitle

\begin{abstract}
    Current neural network-based image classifiers are susceptible to adversarial
examples, even in the black-box setting, where the attacker is limited to query
access without access to gradients. Previous methods --- substitute networks
and coordinate-based finite-difference methods --- are either unreliable or
query-inefficient, making these methods impractical for certain problems.

We introduce a new method for reliably generating adversarial examples under
more restricted, practical black-box threat models. First, we apply natural
evolution strategies to perform black-box attacks using two to three orders of
magnitude fewer queries than previous methods. Second, we introduce a new
algorithm to perform targeted adversarial attacks in the partial-information
setting, where the attacker only has access to a limited number of target
classes. Using these techniques, we successfully perform the first targeted
adversarial attack against a commercially deployed machine learning system, the
Google Cloud Vision API, in the partial information setting.

\end{abstract}

\section{Introduction}
\label{sec:introduction}


Neural network-based image classifiers, despite surpassing human ability on
several benchmark vision tasks, are susceptible
to \textit{adversarial examples}. Adversarial examples are correctly classified images that are minutely perturbed
to cause misclassification. Targeted adversarial examples cause misclassification as a chosen class, while untargeted adversarial examples cause just misclassification.

The existence of these adversarial examples and the feasibility of constructing
them in the real world~\cite{goodfellow-physical,robustadv} points to potential
exploitation, particularly in the face of the rising popularity of neural networks in
real-world systems. For commercial or proprietary systems, however, adversarial
examples must be considered under a much more restrictive threat model. First,
these settings are \textit{black-box}, meaning that an attacker only has access
to input-output pairs of the classifier, often through a binary or API.
Furthermore, often the attacker will only have access to a subset of the
classification outputs (for example, the top $k$ labels and scores); to our knowledge this
setting, which we denote the \textit{partial-information} setting, has not been
considered in prior work.


Prior work considering constrained threat models have only considered the
black-box restriction we describe above; previous work primarily uses
\textit{substitute networks} to emulate the attacked network, and then attack
the substitute with traditional first-order white-box methods~\cite{papernot16,papernot17}. However, as
discussed thoroughly in~\cite{zoo}, this approach is unfavorable for many
reasons including imperfect transferability of attacks from the substitute to the
original model, and the computational and query-wise cost of training a
substitute network. Recent attacks such as~\cite{zoo} have used finite difference methods in order to
estimate gradients in the black-box case, but are still expensive, requiring
millions of queries to generate an adversarial image for an
ImageNet classifier. Effects such as low throughput, high latency, and rate
limiting on commercially deployed black-box classifiers heavily impact the
feasibility of current approaches to black-box attacks on real-world
systems.



We present an approach for generating black-box adversarial
examples based on Natural Evolutionary Strategies~\cite{nes}. We provide
motivation for the algorithm in terms of finite difference
estimation in random Gaussian bases. We demonstrate the effectiveness of
the method in practice, generating adversarial examples with several orders of
magnitude fewer queries compared to existing methods. We consider the
further constrained partial-information setting, and we present a new algorithm for
attacking neural networks under these conditions. We demonstrate the
effectiveness of our method by showing that it can reliably produce
targeted adversarial examples with access to partial input-output pairs.

We use the newfound tractability given by these methods to both (a) generate the
first transformation-tolerant black-box adversarial examples and (b) perform the first targeted
attack on the Google Cloud Vision API, demonstrating the effectiveness of our
proposed method on large, commercial systems: the GCV API is an opaque (no
published enumeration of labels), partial-information (queries return
only up to 10 classes with uninterpretable ``scores''), several-thousand-way
commercially deployed classifier.

Our contributions are as follows:

\begin{itemize}
    \item We propose a variant of NES inspired by the treatement
        in~\cite{nes} as a method for generating black-box
        adversarial examples. We relate NES in this special case with the
        finite difference method over Gaussian bases, providing a theoretical
        comparison with previous attempts at black-box adversarial examples.

    \item We demonstrate that our method is effective in efficiently
        synthesizing adversarial examples; the method does not require
        a substitute network and is 2-3 orders of magnitude
        faster than optimized finite difference-based methods such
        as~\cite{zoo}. We reliably produce black-box adversarial
        examples for both CIFAR-10 and ImageNet classifiers.

    \item We propose an approach for synthesizing targeted adversarial examples
        in the ``partial information'' setting, where the attacker has access
        only to top-$k$ outputs of a classifier, and we demonstrate its
        effectiveness.

    \item We exploit the increased efficiency of this method
        to achieve the following results:

	\begin{itemize}
        \item \textbf{Robust black-box examples.} In~\cite{robustadv}, the
            inability of standard-generated adversarial examples to remain
            adversarial under transformation is noted, and the
            \textit{Expectation over Transformation (EOT)} algorithm is
            introduced. By integrating EOT with the method presented in this
            work, we generate the first transformation-tolerant black-box
            adversarial examples.

	\item \textbf{Targeted adversarial examples against a several-thousand-way commercial classifier.} We use
            our method to generate adversarial examples for the Google Cloud
            Vision API, a commercially-deployed system.
            An attack against a commercial
            classifier of this order of magnitude demonstrates the applicability
            and reliability of our method.
	\end{itemize}

\end{itemize}


\section{Approach}
\label{sec:approach}
We outline the key technical components of our approach allowing us to
attack the constructed threat model. First we describe our application of Natural
Evolutionary Strategies~\cite{nes}. Then, we outline the strategy used
to construct adversarial examples in the \textit{partial-information} setting.

\subsection{Natural Evolutionary Strategies}
\label{sec:nes}
Rather than taking component-wise finite differences as in previous state-of-the
art methods~\cite{zoo}, we use natural evolutionary strategies~\cite{nes}. Natural
evolutionary strategies (NES) is a method for
derivative-free optimisation based on the idea of a \textit{search distribution}
$\pi(\theta|x)$. In particular, rather than maximizing the objective function $F(x)$ directly,
NES maximizes the expected value of the loss function under the
search distribution. As demonstrated in Section~\ref{sec:evaluation}, this
allows for gradient estimation in far fewer queries than typical
finite-difference methods. Concretely, for a loss function $F(\cdot)$ and a current
set of parameters $x$, we have from~\cite{nes}:

\begin{align*}
    \mathbb{E}_{\pi(\theta|x)}\left[F(\theta)\right] &= \int F(\theta)\pi(\theta|x) \diff \theta \\
    \nabla_x \mathbb{E}_{\pi(\theta|x)}\left[F(\theta)\right] &= \nabla_x \int F(\theta)\pi(\theta|x) \diff \theta \\
    &= \int F(\theta) \nabla_x \pi(\theta|x) \diff \theta \\
    &= \int F(\theta) \frac{\pi(\theta|x)}{\pi(\theta|x)} \nabla_x \pi(\theta|x) \diff \theta \\
    &= \int \pi(\theta|x) F(\theta) \nabla_x \log\left(\pi(\theta|x)\right) \diff \theta \\
    &= \mathbb{E}_{\pi(\theta|x)}\left[F(\theta)\nabla_x \log\left(\pi(\theta|x)\right)\right]
\end{align*}

In a manner similar to that in~\cite{nes}, we choose a search distribution
of random Gaussian noise around the current image $x$; that is, we have $\theta
= x + \sigma\mathcal{N}(0, I)$. Evaluating the gradient above with this search
distribution yields the following variance-reduced gradient estimate:

$$\nabla\mathbb{E}[F(\theta)] \approx \frac{1}{\sigma n}\sum_{i=1}^n \epsilon_i
F(\theta + \sigma\epsilon_i)$$

Similarly to~\cite{salimans}, we employ \textit{antithetic sampling} to generate
batches of $\epsilon_i$ values; rather than generating $n$ values $\epsilon_i
\sim \mathcal{N}(0,1)$, we instead draw these values for $i \in \{1\ldots
\frac{n}{2}\}$, and set $\epsilon_j = -\epsilon_{n-j+1}$ for $j \in
\{(\frac{n}{2}+1)\ldots n\}$. This optimization has been empirically
shown to improve performance of NES.

Finally, we perform a projected gradient descent update \cite{madry-adversarial}
with momentum based on the NES gradient estimate.

\subsubsection{NES as Finite Differences}

A closer inspection of the special case of NES that we have described here
suggests an alternative view of the algorithm. In particular, note that when
antithetic sampling is used, the gradient estimate can
be written as the following, where $D_v$ represents the directional derivative
in the direction of $v$:

\begin{align*}
    \nabla \mathbb{E}[F(x)] &\approx \frac{1}{n\sigma}\sum_{i=1}^{n} F(x +
    \sigma\epsilon_i)\epsilon_i \\
    &= \frac{1}{n/2} \sum_{i=1}^{n/2} \frac{F(x + \sigma\epsilon_i) - F(x -
    \sigma\epsilon_i)}{2\sigma}\epsilon_i \\
    &\approx \frac{1}{n/2} \sum_{i=1}^{n/2} D_{\epsilon_i}(x)\epsilon_i \\
    &= \frac{1}{n/2} \sum_{i=1}^{n/2} (\nabla F \cdot \epsilon_i)\epsilon_i
\end{align*}

Now, the $\epsilon_i$ are effectively randomly drawn Gaussian vectors
of size $\textit{width}\cdot\textit{height}\cdot\textit{channels}$. By
a well-known result, these vectors are nearly orthogonal; a formalization of this
is in~\cite{quasiortho}, which says that for an $n$-dimensional space and $N$
randomly sampled Gaussian vectors $v_1 \ldots v_N$,

$$N \leq e^\frac{\delta^2 n}{4}[-\ln(\theta)]^\frac{1}{2} \implies
\mathbb{P}\left\{\frac{v_i\cdot v_j}{||v_i||||v_j||} \leq \delta\ \forall\ (i,
j)\right\} = \theta$$

Thus, one can ``extend'' the randomly sampled vectors into a complete basis of
the space $[0,1]^n$; then we can perform a basis decomposition on $\nabla F(x)$
to write:
$$\nabla F(x) = \sum_{i=1}^{n} \langle \nabla F, \epsilon_i \rangle \epsilon_i$$

Thus, the NES gradient can be seen as essentially ``clipping'' this space to
the first $N$ Gaussian basis vectors and performing a finite-differences
estimate.

More concretely, considering a matrix $\Theta$ with $\epsilon_i$
being the columns and the projection $\Theta (\nabla F)$, we can use
results from concentration theory to analyze our estimate, either through the
following simple canonical bound or a more complex treatment such as is
given in~\cite{mixedgaussian}:

$$\mathbb{P}\left\{(1-\epsilon)||\nabla||^2 \leq ||\Theta \nabla||^2 \leq
(1+\epsilon)||\nabla||^2\right\} \geq 1 - 2e^{-c\epsilon^2 m}$$

Note that even more rigorous analyses of such ``Gaussian-projected finite
difference'' gradient estimates and bounds have been demonstrated by works such
as~\cite{nesterov}, which detail the algorithm's interaction
with dimensionality, scaling, and various other factors.

\subsection{Partial-Information Setting}
Next, we consider the partial-information setting described in the previous
section. In particular, we now assume access to both probabilities and  gradient
approximations through the methods described in~Section~\ref{sec:nes}, but only for the
top $k$ classes $\{y_1,\ldots,y_k\}$. In normal settings, given an image and
label $(x_i,y)$, generating an adversarial example $(x_{adv}, y_{adv})$ for a
targeted $y_{adv}$ can be acheived using standard first-order attacks. These are attacks which
involve essentially ascending the estimated gradient $\nabla P(y_{adv}|x)$. However,
in this case $P(y_{adv}|x_i)$ (and by extension, its gradient) is unavailable to the
classifier.

To resolve this, we propose the following algorithm. Rather than beginning with
the image $x_i$, we instead begin with an image $x_0$ \textit{of the original
target class}. Then $y_{adv}$ will be in the top-$k$ classes for
$x_0$. We perform the following iterated optimization:

\begin{align*}
    \epsilon_t &= \min \epsilon \text{  s.t.
    $\text{rank}\left(P\left(y_{adv}|\Pi_\epsilon(x_{t-1})\right)\right) < k$} \\
    x_t &= \arg\max_{x} P(y_{adv}|\Pi_{\epsilon_{t-1}}(x))
\end{align*}

where $\Pi_\epsilon(x)$ represents the $\ell_\infty$ projection of $x$ onto the
$\epsilon$-box of $x_i$. In particular, we concurrently perturb the image to
maximize its adversarial probability, while projecting onto $\ell_\infty$ boxes
of decreasing sizes centered at the original image $x_i$, maintaining that the
adversarial class remains within the top-$k$ at all times. In practice, we
implement this iterated optimization using backtracking line search to find
$\epsilon_t$, and several iterations projected gradient descent (PGD) to find
$x_t$. Alternatingly updating $x$ and $\epsilon$ until $\epsilon$ reaches the
desired value yields an adversarial example that is $\epsilon$-away
from $x_i$ while maintaining the adversarial classification of the original
image.

\section{Threat model}
\label{sec:threat_model}

Our threat model is chosen to model the constraints of attacking deep neural
networks deployed in the real world.

\begin{itemize}

    \item \textbf{No access to gradients, logits, or other internals.} Similar
        to previous work, we define \textit{black-box} to mean that access to
        gradients, logits, and other network internals is unavailable.
        Furthermore, the attacker does not have knowledge of the network
        architecture. The attacker only has access to the output of the
        classifier: prediction probabilities for each class.

    \item \textbf{No access to training-time information.} The attacker has no
        information about how the model was trained, and the attacker does not
        have access to the training set.

    \item \textbf{Limited number of queries.} In real-time models like
        self-driving cars, the format of the input allows us to make a large
        number of queries to the network (e.g. by disassembling the car,
        overriding the input signals, and measuring the output signals). In most
        other cases, proprietary ML models like the Google Cloud Vision API
        are rate-limited or simply unable to support a large number of queries
        to generate a single adversarial example.

    \item \textbf{Partial-information setting:} As discussed in
	Section~\ref{sec:introduction}, we also consider in this work the case where
	the full output of the classifier is unavailable to the attacker. This
	more accurately reflects the state of commercial systems where even the list of possible classes is unknown to the attacker, such as in the Google Cloud Vision API, Amazon's Rekognition API, or the Clarifai API.

\end{itemize}

Attackers can have one of two goals: untargeted or targeted misclassification,
where targeted attacks are strictly harder. A successful targeted adversarial
example is one that is classified as a specific target class. An untargeted
adversarial example is one that is misclassified.

Notably, we omit wall-clock time to attack as a security parameter in our
threat model. This metric is more indicative of hardware resources used for the
attack than the efficacy of an attack itself, for which query count is a
realistic and practical measure.

\section{Evaluation}
\label{sec:evaluation}

\subsection{Targeted black-box adversarial examples}

We evaluate the effectiveness of our black-box attack in generating targeted
adversarial examples for neural networks trained on CIFAR-10 and ImageNet.
We demonstrate our attack against the CIFAR-10 network of Carlini
and Wagner~\cite{sp2017:carlini} and the InceptionV3
network~\cite{szegedy-inception} in the black-box setting, assuming
access to the output probabilities of the classifiers. For each of the
classifiers, we randomly choose 1000 examples from the test set, and for each
example, we choose a random target class. We then use projected gradient
descent (PGD)~\cite{madry-adversarial} with NES gradient estimates, maximizing
the log probability of the target class while constraining to a maximum
$\ell_\infty$ perturbation of $\epsilon = 0.05$. We use a fixed set of
hyperparameters across all attacks on a single classifier, and we run the
attack until we produce an adversarial image or until we time out (at a maximum
of 1 million queries).

\begin{figure}
	\begin{centering}
        \includegraphics[width=\linewidth]{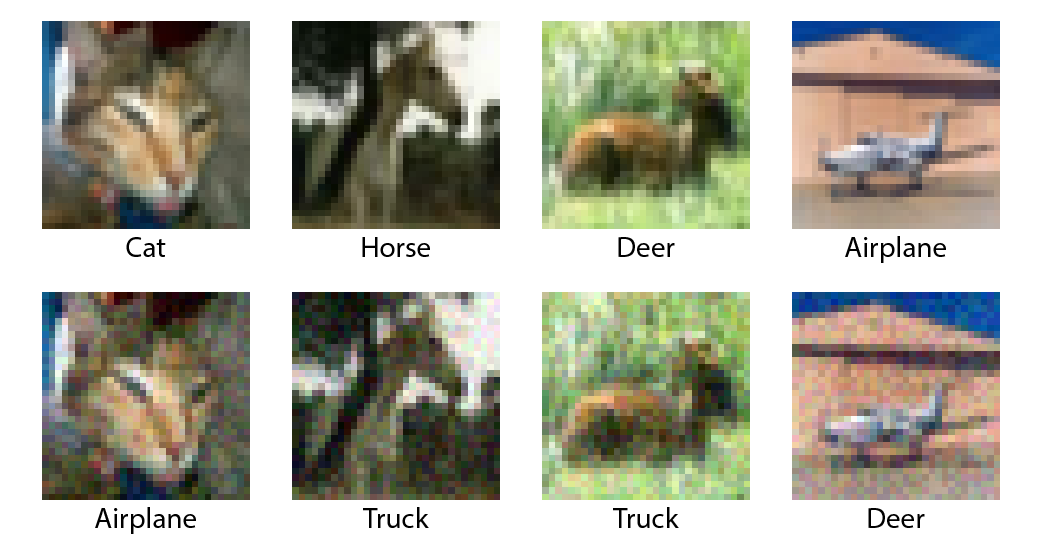}
        \caption{
        Randomly chosen samples from the 1000 adversarial examples for the CIFAR-10 network.
        }
		\label{fig:single-viewpoint-cifar}
        \includegraphics[width=\linewidth]{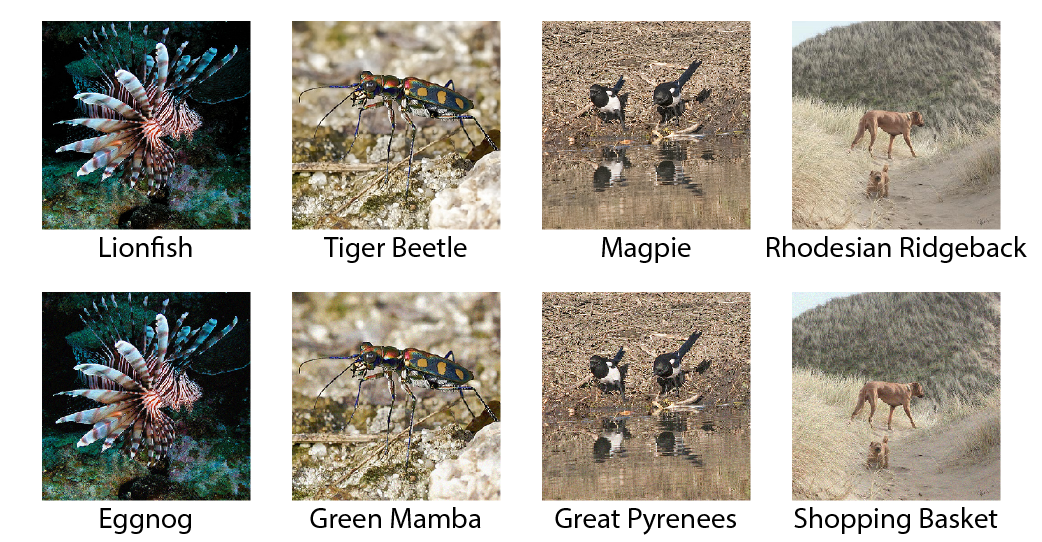}
        \caption{
        Randomly chosen samples from the 1000 adversarial examples for the InceptionV3 network.
        }
		\label{fig:single-viewpoint-imagenet}
    \end{centering}
\end{figure}

\begin{figure}
	\begin{centering}
        \includegraphics[width=\linewidth]{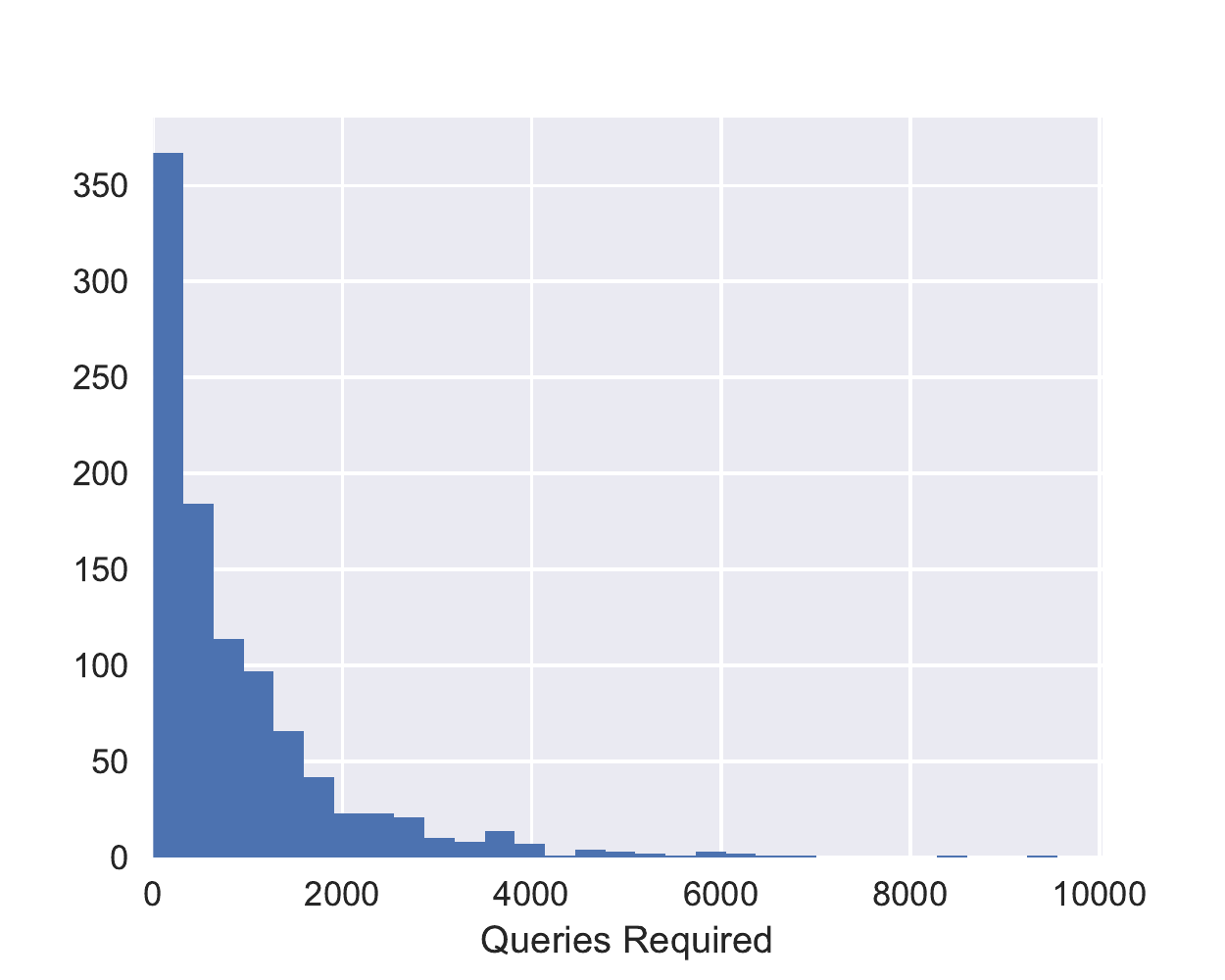}
        \caption{
            Distribution of number of queries required to generate an adversarial
            image with a randomly chosen target class for the
            \textbf{CIFAR-10} network over the 1000 test images.
        }
		\label{fig:cifar-histogram}
    \end{centering}
\end{figure}

\begin{figure}
	\begin{centering}
        \includegraphics[width=\linewidth]{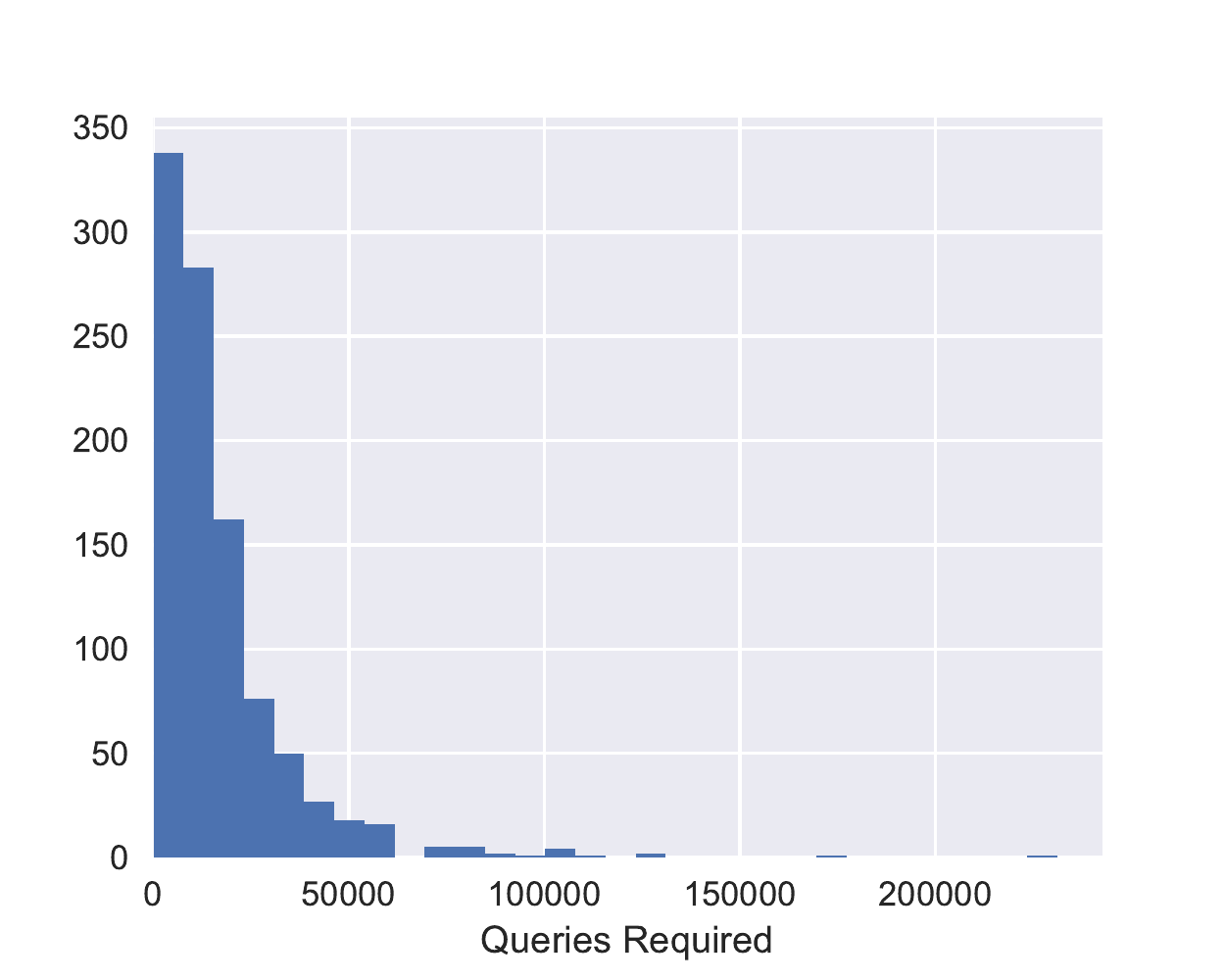}
        \caption{
            Distribution of number of queries required to generate an adversarial
            image with a randomly chosen target class for the
            \textbf{InceptionV3 ImageNet} network over the 1000 test
            images.
        }
		\label{fig:imagenet-histogram}
    \end{centering}
\end{figure}

Table~\ref{tab:single-viewpoint} summarizes the results of our experiment. Our
attack is highly effective and query-efficient, with a \textbf{99.6\%} success
rate on CIFAR-10 with a mean of 4910 queries to the black-box classifier
per example, and a \textbf{99.2\%} success rate on ImageNet with a mean of
24780 queries to the black-box classifier per example.
Figures~\ref{fig:single-viewpoint-cifar}~and~\ref{fig:single-viewpoint-imagenet}
show a sample of the adversarial examples we produced.
Figures~\ref{fig:cifar-histogram}~and~\ref{fig:imagenet-histogram} show the
distribution of number of queries required to produce an adversarial example:
in most cases, the attack requires only a small number of queries.

\begin{table*}
	\begin{centering}
        \begin{tabular}{c c c c c}
        \toprule
            \textbf{Dataset} & \phantom{abc} & \textbf{Original Top-1 Accuracy} & \textbf{Attack Success Rate} & \textbf{Mean Queries} \\
            \midrule
            CIFAR-10 && 80.5\% & \textbf{99.6\%} & 4910 \\
            ImageNet && 77.2\% & \textbf{99.2\%} & 24780 \\
            \bottomrule
        \end{tabular}
        \caption{
            Quantitative analysis of targeted adversarial attacks we perform on 1000 randomly chosen test images and randomly chosen target classes. The attacks are limited to 1 million queries per image, and the adversarial perturbations are constrained with $\l_{\infty}, \epsilon=.05$. The same hyperparameters were used for all images in each dataset.
        }
		\label{tab:single-viewpoint}
    \end{centering}
\end{table*}

\subsection{Robust black-box adversarial examples}
We evaluate the effectiveness of our black-box attack in generating
adversarial examples that fool classifiers over a distribution of
transformations using Expectation-Over-Transformation
(EOT)~\cite{robustadv}. In this task, given a distribution of
transformations $T$ and $\ell_\infty$ constraint $\epsilon$, we
attempt to find the adversarial example $x'$ (for some original input
$x$) that maximizes the classifier's expected output probability of a
target class $y$ over the distribution of inputs $T(x')$:
\[
  x' = \argmax\limits_{x', \|x-x'\|_\infty \leq \epsilon} \mathbb{E}_{t \sim T}[\log P(y|t(x'))]
\]
We use the PGD attack of \cite{madry-adversarial} to solve the EOT
optimization problem, using NES to estimate the gradient of the classifier. Note that $P(y|\cdot)$ is the classifier's
output probability for label $y$ given an input. In our evaluation we
randomly choose 10 examples from the ImageNet validation set, and for each
example we randomly choose a target class. We choose our distribution of
transformations to be a $\theta$ degree rotation
for $\theta \sim \text{unif}(-30\degree,30\degree)$,
and set $\epsilon=0.1$. We use a fixed set of hyperparameters across
all attacks, and perform the PGD attack until we achieve greater than
90\% adversariality on a random sample of 100 transformations.

\begin{figure*}
    \begin{centering}
        \begingroup\renewcommand*{\arraystretch}{3}
\begin{tabular}{C{0.17500\linewidth}|C{0.15625\linewidth}C{0.15625\linewidth}C{0.15625\linewidth}C{0.15625\linewidth}}
\includegraphics[align=c,width=\linewidth]{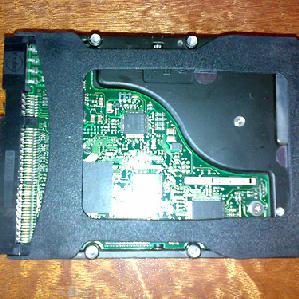} \newline Original: hard disc & \includegraphics[align=c,width=\linewidth]{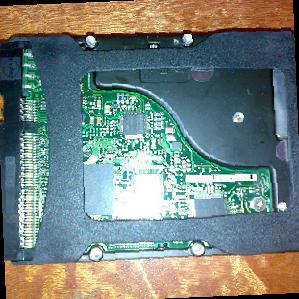} \newline $P(true)$: 100\% \newline $P(adv)$: 0\% & \includegraphics[align=c,width=\linewidth]{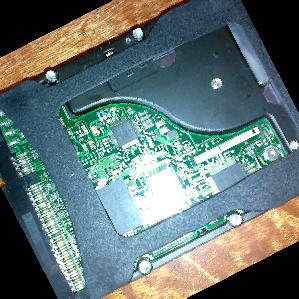} \newline $P(true)$: 100\% \newline $P(adv)$: 0\% & \includegraphics[align=c,width=\linewidth]{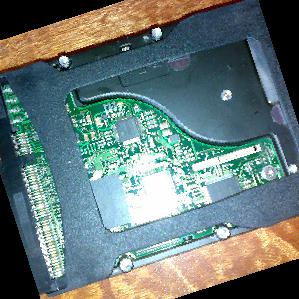} \newline $P(true)$: 100\% \newline $P(adv)$: 0\% & \includegraphics[align=c,width=\linewidth]{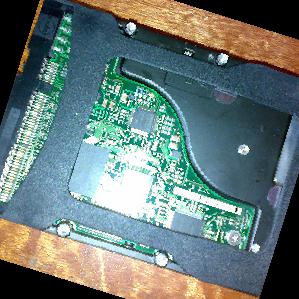} \newline $P(true)$: 100\% \newline $P(adv)$: 0\% \\ 
\includegraphics[align=c,width=\linewidth]{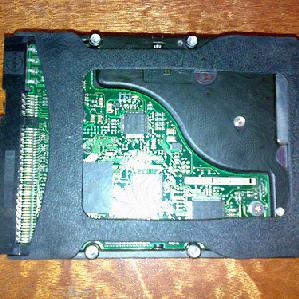} \newline Adversarial: bearskin& \includegraphics[align=c,width=\linewidth]{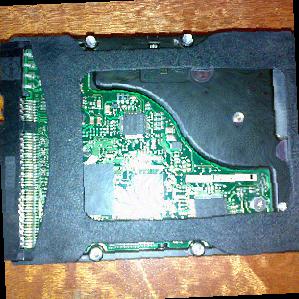} \newline $P(true)$: 0\% \newline $P(adv)$: 70\% & \includegraphics[align=c,width=\linewidth]{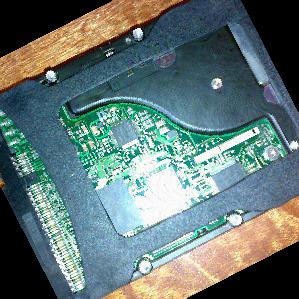} \newline $P(true)$: 3\% \newline $P(adv)$: 45\% & \includegraphics[align=c,width=\linewidth]{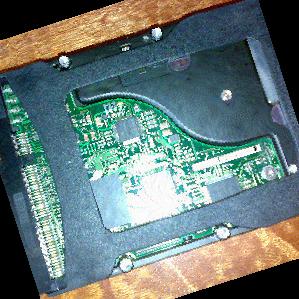} \newline $P(true)$: 1\% \newline $P(adv)$: 41\% & \includegraphics[align=c,width=\linewidth]{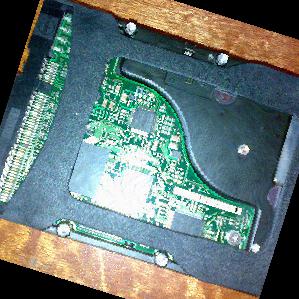} \newline $P(true)$: 1\% \newline $P(adv)$: 62\% \\ 
\end{tabular}
\endgroup

        \caption{
            Random sample of adversarial examples robust up to $\pm
        30\degree$ rotations along with classifier probabilities of
        the adversarial and natural classes.
    }
        \label{fig:multi-viewpoint-sample}
    \end{centering}
\end{figure*}

Table~\ref{tab:multi-viewpoint} shows the results of our
experiment. We achieve a mean attack success rate (where attack success rate
is defined for a single adversarial example as the percentage of randomly
transformed samples that classify as the target class) of
\textbf{95.7\%} on our 10 attacks, and use a mean of
3780000 queries per
example. Figure~\ref{fig:multi-viewpoint-sample} shows samples of
the adversarial examples robust up to $\pm 30\degree$.

\begin{table*}
    \begin{centering}
        \begin{tabular}{c c c c}
            \toprule
            \textbf{Original Top-1 Accuracy} & \textbf{Mean Attack Success Rate} & \textbf{Mean Queries} \\
            \midrule
            80.0\% & \textbf{95.7\%} & 3780000 \\
            \bottomrule
        \end{tabular}
        \caption{
            Mean attack success rate and mean required queries across
          the 10 generated adversarial examples on InceptionV3 robust up to $\pm
          30\degree$ rotations.
      }
        \label{tab:multi-viewpoint}
    \end{centering}
\end{table*}

\subsection{Targeted partial-information adversarial examples}

We evaluate the effectiveness of our partial-information black-box attack in
generating targeted adversarial examples for the InceptionV3 network when given
access to only the top 10 class probabilities out of the total of 1000 labels.
We randomly choose 1000 examples from the test set, and for each example, we
choose a random target class. For each source-target pair, we find an example
of the target class in the test set, initialize with that image, and use our
partial-information attack to construct a targeted adversarial example. We use
PGD with NES gradient estimates, constraining to a maximum $\ell_\infty$
perturbation of $\epsilon = 0.05$. We use a fixed set of hyperparameters across
all attacks, and we run the attack until we produce an adversarial example or
until we time out (at a maximum of 1 million queries).

The targeted partial-information attack achieves a \textbf{95.5\%}
success rate with a mean of 104342 queries to the black-box classifier.

\subsection{Attacking Google Cloud Vision}
\label{sec:gcv}
In order to demonstrate the relevance and applicability of our approach to
real-world system, we attack the Google Cloud Vision API, a commercially
available computer vision suite offered by Google. In particular, we attack the
most general \textit{object labeling} classifier, which performs n-way
classification on any given image. This case is considerably more challenging
than even the typical black-box setting. The number of classes is large and
unknown --- a full enumeration of labels is unavailable. The
classifier returns ``confidence scores'' for each label it assigns to an image,
which seem to be neither probabilities nor logits. The classifier
does not return scores for all labels, but instead returns an
unspecified-length list of labels that varies based on image. Despite these
challenges, we successfully demonstrate the ability of the system to generate
black-box adversarial examples, in both an untargeted attack and a targeted
attack.

\subsubsection{Untargeted attack}

\begin{figure}
    \centering
    \includegraphics[width=\linewidth]{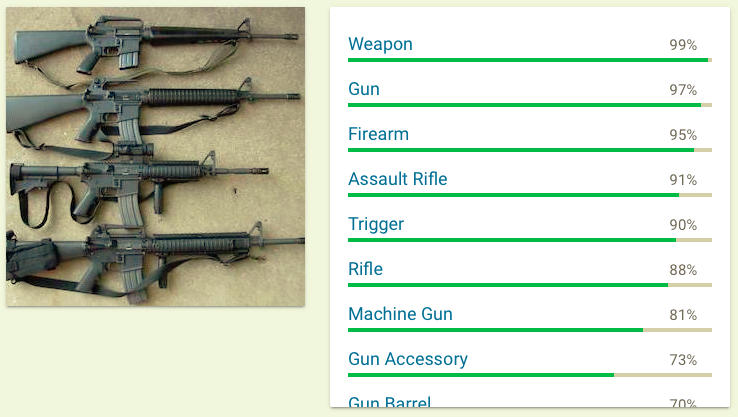}
    \caption{The Google Cloud Vision Demo labelling on the unperturbed image.}
    \label{fig:gcv}
    \includegraphics[width=\linewidth]{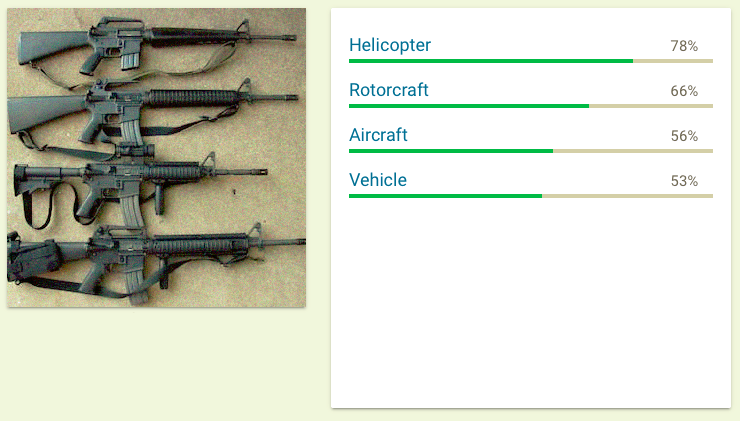}
    \caption{The Google Cloud Vision Demo labelling on the adversarial image
	generated with $\ell_\infty$ bounded perturbation with $\epsilon = 0.1$:
	the original class is no longer a returned label.}
    \label{fig:gcv_adv}
\end{figure}

Figure~\ref{fig:gcv} shows an unperturbed image being correctly labeled as
several rifle/rifle-related classes, including ``weapon'' and ``firearm.'' We run
the algorithm presented in this work, but rather than maximizing the
probability of a target class, we write the following loss function based on
the classification $C(x)$ to minimize the maximum score assigned to any label
semantically similar to ``gun'': $$F^*(x) = \max_{C(x)\text{[``label'']} \sim
``gun''} C(x)\text{[``score'']}$$

Note that we expand the definition of ``misclassification'' to encompass
semantic similarity---that is, we are uninterested in a modification that
induces a classification of ``persian cat'' on a ``tabby cat.'' Applying the
presented algorithm to this loss function with $\epsilon=0.1$ yields the
adversarial example shown in Figure~\ref{fig:gcv_adv}, definitively
demonstrating the applicability of our method to real-world commercial systems.

\subsubsection{Targeted attack}

\begin{figure}
    \centering
    \includegraphics[width=\linewidth]{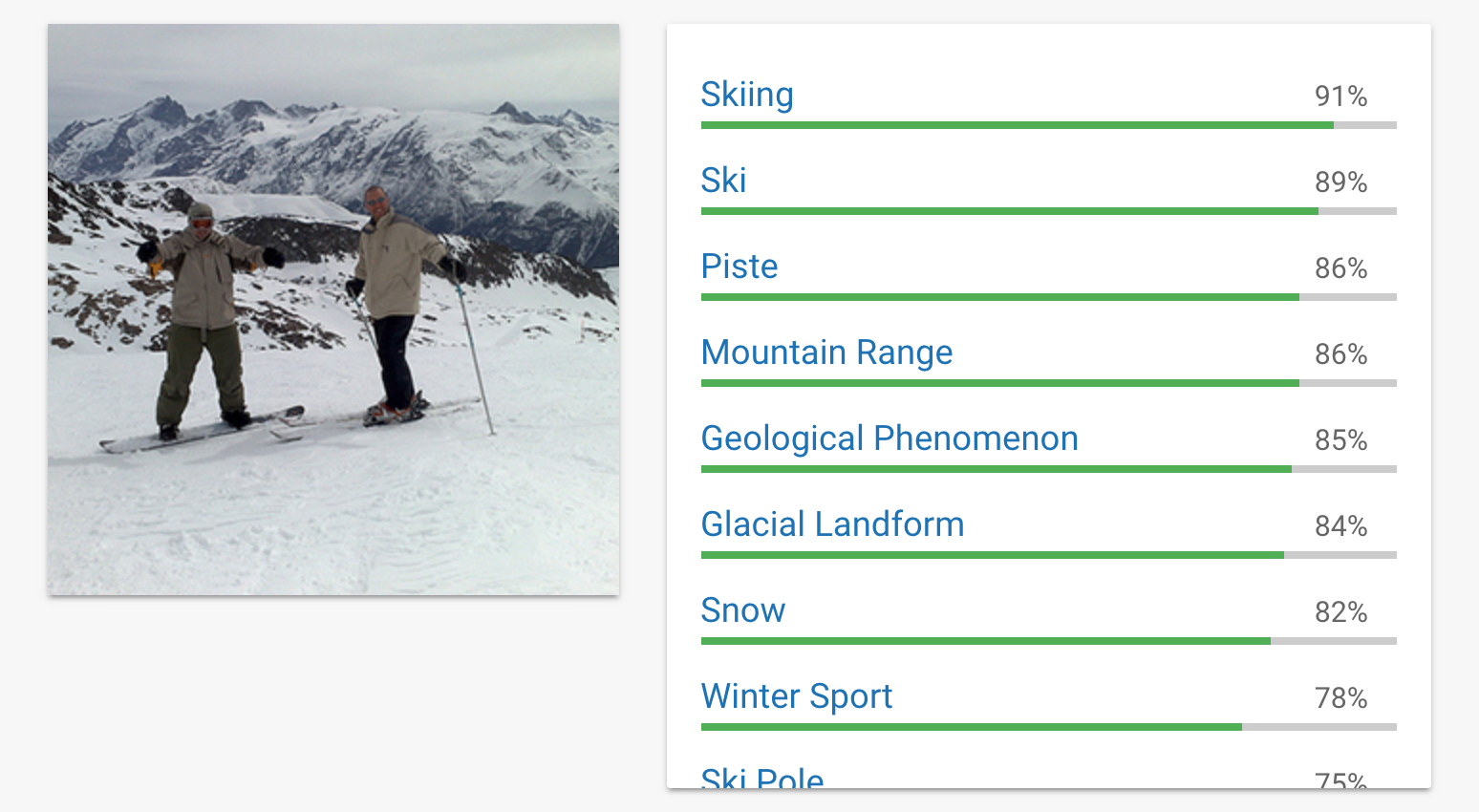}
    \caption{The Google Cloud Vision Demo labelling on the unperturbed image.}
    \label{fig:gcv-targeted}
    \includegraphics[width=\linewidth]{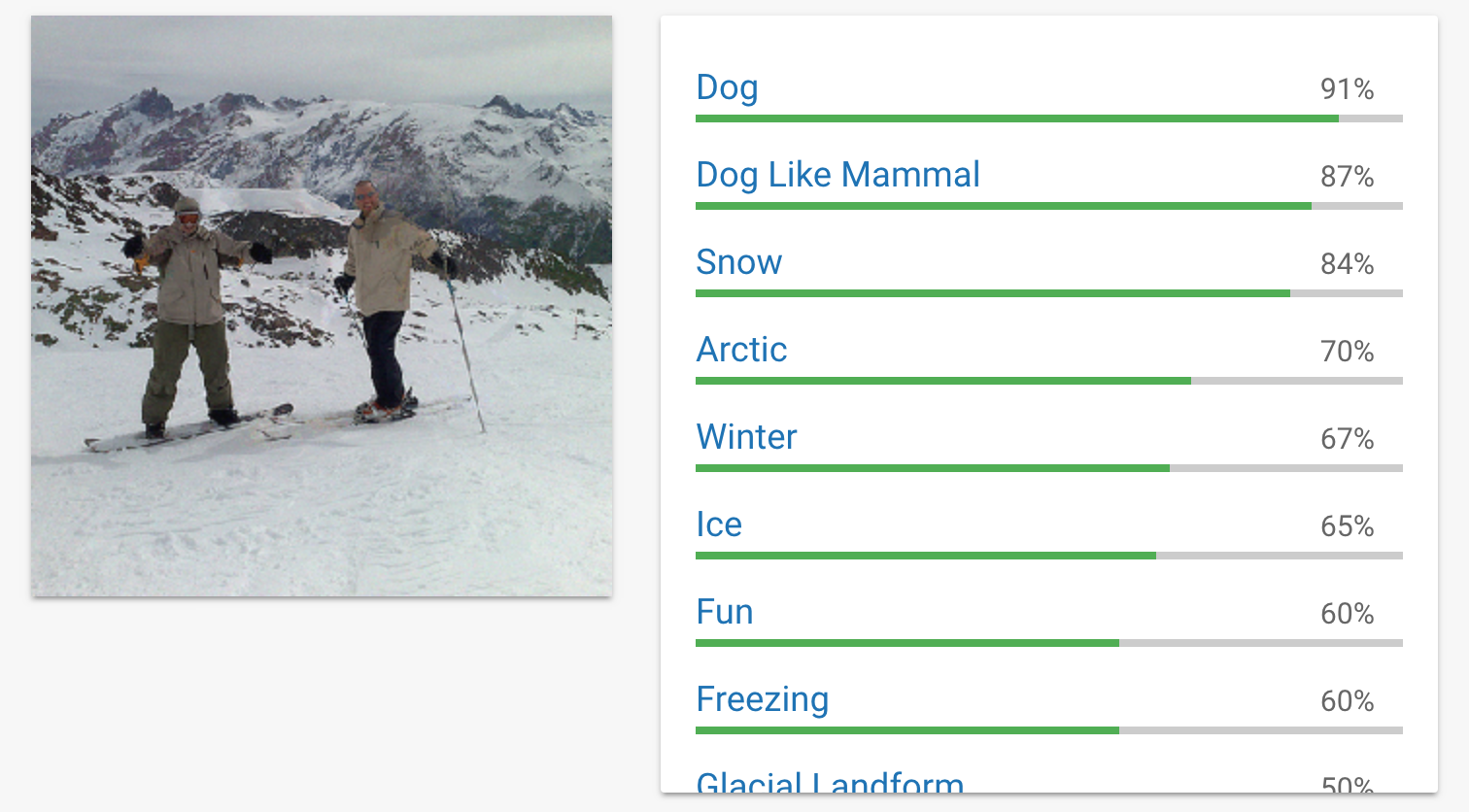}
    \caption{The Google Cloud Vision Demo labelling on the adversarial image
	generated with $\ell_\infty$ bounded perturbation with $\epsilon = 0.1$:
	the image is labeled as the target class.}
    \label{fig:gcv-targeted-adv}
\end{figure}

Figure~\ref{fig:gcv-targeted} shows an unperturbed image being correctly
labeled as several skiing-related classes, including ``skiing'' and ``ski''. We
run our partial-information attack to force this image to be classified as
``dog''. Note that the label ``dog'' does not appear in the output for the
unperturbed image. We initialize the algorithm with a photograph of a dog
(classified by GCV as a dog) and use our partial-information attack to
synthesize an image that looks like the skiers but is classified as ``dog''.

\section{Related work}
\label{sec:related-work}

Szegedy et al. (2014)~\cite{szegedy} first demonstrated that
neural networks are vulnerable to adversarial examples.
A number of techniques have been developed to generate
adversarial examples in the white-box case~\cite{iclr2015:goodfellow,sp2017:carlini,goodfellow-physical}, where an attacker is
assumed to have full access to the model parameters and architecture.

Previous work has shown that adversarial examples can be
generated in the black-box case by training a substitute model,
and then exploiting white-box techniques on the substitute~\cite{papernot16,papernot17}. However, this attack is unreliable because
it assumes the adversarial examples can transfer to the target model.
At best, the adversarial images are less effective, and in some cases,
attacks may entirely fail to transfer and ensembles of substitute models
need to be used~\cite{iclr2017:liu}. Our attack does not
require the transferability assumption.

Recent attempts use finite differences to estimate the gradient instead
of training substitute models~\cite{zoo}. However, even with various query
reduction techniques, a large number of queries are required to
generate a single attack image, potentially rendering
real-world attacks and transformation-tolerant adversarial examples intractable.
In comparison, our method uses several orders of magnitude fewer queries.

Prior work has demonstrated that black-box methods can feasibly attack
real-world, commercially deployed systems, including image classification APIs
from Clarifai, Metamind, Google, and
Amazon~\cite{iclr2017:liu,papernot17,hayes}, and a speech recognition system
from Google~\cite{usenix16:carlini}. Our work advances prior work on machine
learning systems deployed in the real world by demonstrating a highly effective
and query-efficient attack against the Google Cloud Vision API in the
partial-information setting, a scenario that has not been explored in prior work.

\section{Conclusion}
\label{sec:conclusion}

In this work, we present an algorithm based on natural evolutionary strategies
(NES) which allows for the generation of adversarial examples in the black-box
setting without training a substitute network. We also introduce the
partial-information setting, a more restricted black-box situation that better
models large-scale commercial systems, and we present an algorithm for crafting
targeted adversarial examples for this setting. We motivate our algorithm
through the formulation of NES as a set of finite differences over a random
normal projection, and demonstrate the empirical efficacy of the method by
generating black-box adversarial examples orders of magnitude more efficient
(in terms of number of queries) than previous work on both the CIFAR-10 and
ImageNet datasets. Using a combination of the described algorithm and the EOT
algorithm, we generate the first robust black-box adversarial examples, which
constitutes a step towards attacking real-world systems. We also demonstrate
the efficacy of our partial-information attack. Finally, we synthesize targeted
adversarial examples for the commercial Google Cloud Vision API, demonstrating
the first targeted attack against a partial-information system. Our results
point to a promising new method for efficiently and reliably generating
black-box adversarial examples.

\ifcvprfinal
    \subsection*{Acknowledgements}

Special thanks to Nat Friedman and Daniel Gross.

\fi
{\small
\bibliographystyle{ieee}
\bibliography{paper}
}

\end{document}